\documentclass[10pt, a4paper]{article}
\usepackage{lrec}
\usepackage{multibib}
\usepackage{graphicx}
\usepackage{tabularx}
\usepackage{dirtytalk}
\usepackage{setspace}
\usepackage{stackengine}
\usepackage{algorithm}
\usepackage{algpseudocode}
\usepackage{booktabs,lipsum}
\usepackage{soul}
\usepackage{amsmath}

\usepackage{epstopdf}
\usepackage[latin1]{inputenc}

\usepackage{hyperref}
\usepackage{xstring}
\makeatletter
\def\BState{\State\hskip-\ALG@thistlm}
\makeatother

\title{WikiRank:Improving Keyphrase Extraction Based on Background Knowledge \\ \vspace*{.5\baselineskip}}

\name{Yang Yu, Vincent Ng}

\address{Human Language Technology Research Institute \\ University of Texas at Dallas\\ \{yangyu,vince\}@hlt.utdallas.edu \\}

\abstract{Keyphrase is an efficient representation of the main idea of documents. While background knowledge can provide valuable information about documents, they are rarely incorporated in keyphrase extraction methods. In this paper, we propose WikiRank, an unsupervised method for keyphrase extraction based on the background knowledge from Wikipedia.  Firstly, we construct a semantic graph for the document. Then we transform the keyphrase extraction problem into an optimization problem on the graph. Finally, we get the optimal keyphrase set to be the output. Our method obtains improvements over other state-of-art models by more than 2\% in F1-score. \\\newline \Keywords{Keyphrase Extraction, Knowledge Graph, Semantic Graph} }

\begin{document}
\maketitleabstract

\section{Introduction}
As the amount of published material rapidly increases, the problem of managing information becomes more difficult. Keyphrase, as a concise representation of the main idea of the text, facilitates the management, categorization, and retrieval of information. Automatic keyphrase extraction concerns ``the automatic selection of important and topical phrases from the body of a document''. Its goal is to extract a set of phrases that are related to the main topics discussed in a given document\cite{hasan-ng:2014:P14-1}.\\
Existing methods of keyphrase extraction could be divided into two categories: supervised and unsupervised. While supervised approaches require human labeling, at the same time needs various kinds of training data to get better generalization performance, more and more researchers focus on unsupervised methods.\\
Traditional methods of unsupervised keyphrase extraction mostly focus on getting information of document from word frequency and document structure\cite{hasan-ng:2014:P14-1}, however, after years of attempting,  the performance seems very hard to be improved any more. Based on this observation, it is reasonable to suspect that the document itself possibly cannot provide enough information for keyphrase extraction task.\\
To get good coverage of the main topics of the document, Topical PageRank \cite{Liu:2010:AKE:1870658.1870694} started to adopt topical information in automatic keyphrase extraction. The main idea of Topical PageRank is to extract the top topics of the document using LDA, then sum over the scores of a candidate phrase under each topic to be the final score. The main problems with Topical PageRank are: First, The topics are too general. Second, since they are using LDA, they only classify the words to several topics, but don't know what the topics exactly are. However, the topical information we need for keyphrase extraction should be precise. As shown in Figure \ref{example}, the difference between a correct keyphrase \textit{sheep disease} and an incorrect keyphrase \textit{incurable disease} could be small, which is hard to be captured by rough topical categorization approach.\\
To overcome the limitations of aforementioned approaches, we propose WikiRank, an unsupervised automatic keyphrase extraction approach that links semantic meaning to text

The key contribution of this paper could be summarized as follows:
\begin{enumerate}
\item We leverage the topical information in knowledge bases to improve the performance of keyphrase extraction.
\item We model the keyphrase extraction as an optimization problem, and provide the corresponding solution as well as a pruning approach to reduce the complexity.
\end{enumerate}

\section{Existing Error Illustration with Example}
Figure \ref{example} shows part of an example document\footnote{Document from DUC-2001 Dataset \textit{AP900322-0200} \textit{Government Boosts Spending to Combat Cattle Plague}}. In this figure, the gold keyphrases are marked with bold, and the keyphrases extracted by the TextRank system are marked with parentheses. We are going to illustrate the errors exist in most of present keyphrase extraction systems using this example.

\newlength\lunderset
\lunderset= -3.0pt\relax
\def\stackalignment{l}
\newcommand\nundertext[2][1]{\def\useanchorwidth{T}\def\stackalignment{l}\smash{%
  \stackunder[#1\lunderset]{}{%
  \scriptsize\strut#2}}}
\newcommand{\textoverline}[1]{$\overline{\mbox{#1}}$}

\begin{figure*}[ht]
\fbox{\noindent\begin{minipage}[t]{\textwidth}%
\setstretch{1.9}
\hspace{0.05pt}(\nundertext{\textoverline{wiki:Bovine\_spongiform\_}encephalopathy}\textbf{Mad cow disease}) \hspace{10pt} has \hspace{10pt} \nundertext{\textoverline{wiki:De}ath}killed 10,000 \nundertext{\textoverline{wiki:Ca}ttle}cattle\hspace{1pt}, restricted the \nundertext{\textoverline{wiki:Expo}rt}\textbf{export}\hspace{5pt} \nundertext{\textoverline{wiki:Mark}et\_(economics)}market\hspace{25pt} for\hspace{5pt} \nundertext{\textoverline{wiki:Briti}sh\_Empire}Britain's\hspace{20pt} (\nundertext{\textoverline{wiki:Agribusiness\hspace{6pt}}}cattle industry) and raised \nundertext{\textoverline{wiki:Fear}}fears about the \nundertext{\textoverline{wiki:Saf}ety}safety of \nundertext{\textoverline{wiki:Eat}ing}eating\hspace{7pt} \nundertext{\textoverline{\hspace{2pt}wiki:\hspace{1pt}}Beef}beef. The \nundertext{\textoverline{wiki:Government\hspace{1pt}}}\textbf{government} insists the \nundertext{\textoverline{wiki:Disea}se}disease\hspace{1pt} poses only a remote \nundertext{\textoverline{wiki:}Risk}risk to human \nundertext{\textoverline{wiki:Hea}lth}health\hspace{1pt}, but \nundertext{\textoverline{wiki:Scientis}t}scientists still aren't certain what \nundertext{\textoverline{wiki:Cau}sality}causes the disease or h\nundertext{wiki:Transmission (medicine)}ow it is \underline{transmitted} \nundertext{wiki:Bovine\_spongiform\_encephalopathy}\ldots\hspace{0.05pt}(\hspace{1pt}\underline{\textbf{Mad cow disease}})\hspace{2pt},\hspace{10pt} or\hspace{10pt} \nundertext{\textoverline{wiki:Bovine\_spongiform\_encephalopathy\hspace{30pt}}}Bovine Spongiform Encephalopathy, or \nundertext{\textoverline{wiki:B}ovine\_spongiform\_encephalopathy}\textbf{BSE}, was diagnosed only in 1986. The \nundertext{\textoverline{wiki:Symptom}}symptoms are very much like \nundertext{\textoverline{wiki:Scrap}ie}\textbf{scrapie}\hspace{2pt}, a (\nundertext{\textoverline{wiki:She}ep}\textbf{sheep disease}) which has been in \nundertext{\textoverline{wiki:Grea}t Britain}Britain since the 1700s. The (\nundertext{\textoverline{wiki:Cure\hspace{9pt}}}incurable disease) \nundertext{\textoverline{wiki:}Cannibalism}eats holes in the \nundertext{\textoverline{\hspace{2pt}wiki:Hu}man\_Brain}brains of its victims; in late stag\nundertext{wiki:Disease}es a \underline{sick} \nundertext{wiki:Animal}\underline{animal} may act skittish or stagger drunkenly \vspace{10pt}\ldots The \nundertext{\textoverline{wiki:Government}}\textbf{government} \nundertext{wiki:Ban (law)}\underline{banned} the use of sheep \nundertext{wiki:Offal}\underline{offal} in (cattle feed) in June 1988, and later banned the use of (cattle\nundertext{wiki:Brain} \underline{brain}), \nundertext{\textoverline{wiki:Sple}en}spleen \ldots has propos\nundertext{wiki:Ban (law)}ed a \underline{ban} on\nundertext{w\textoverline{iki:Export}} \textbf{exports}\nundertext{wiki:United Kingdom}\hspace{5pt} of\hspace{5pt} (\underline{\textbf{British}}\hspace{20pt} \nundertext{wiki:Cattle}\underline{\textbf{cattle}}) older than 6 months\nundertext{wiki:Bovine spongiform encephalopathy} \ldots has complained of ``\underline{\textbf{BSE}} \nundertext{\textoverline{wiki:Mass }hysteria}hysteria'' in the \nundertext{wiki:Mass media}\underline{media} and has insisted that the \nundertext{wiki:Risk}\underline{risk} of the \nundertext{wiki:Disease}(\underline{disease} passing) to \nundertext{wiki:Human}\underline{humans} is ``remote.'' \ldots known as (\nundertext{wiki:Creutzfeldt-Jakob disease}\underline{Creutzfeldt Jakob disease}). About two dozen cases were reporte\nundertext{wiki:Great Britain}d in (\underline{Britain} last year).   

\end{minipage}
}

\caption[]
    {\tabular[t]{@{}l@{}} Part of the Sample Document \setcounter{footnote}{1}\footnotemark\\\scriptsize Bold: Gold Keyphrase \hspace{10pt}In parentheses: Keyphrase generated by TextRank algorithm \hspace{10pt} Underlined: Keyphrase annotated to Wikipedia Entity by TagMe\endtabular}
\label{example}
\end{figure*}

\footnotetext{Prefix ``wiki'' represents the namespace ``https://en.wikipedia.org/wiki/''}

\textbf{Overgeneration errors} occur when a system correctly predicts a candidate as a keyphrase because it contains a word that frequently appears in the associated document, but at the same time erroneously outputs other candidates as keyphrases because they contain the same word\cite{hasan-ng:2014:P14-1}. It is not easy to reject a non-keyphrase containing a word with a high term frequency: many unsupervised systems score a candidate by summing the score of each of its component words, and many supervised systems use unigrams as features to represent a candidate. To be more concrete, consider the news article in Figure \ref{example}. The word Cattle has a significant presence in the document. Consequently, the system not only correctly predict \textit{British cattle} as a keyphrase, but also erroneously predict \textit{cattle industry}, \textit{cattle feed}, and \textit{cattle brain} as keyphrases, yielding overgeneration errors.\\
\textbf{Redundancy errors} occur when a system correctly identifies a candidate as a keyphrase, but at the same time outputs a semantically equivalent candidate (e.g., its alias) as a keyphrase. This type of error can be attributed to the failure of a system to determine that two candidates are semantically equivalent. Nevertheless, some researchers may argue that a system should not be penalized for redundancy errors because the extracted candidates are in fact keyphrases. In our example, \textit{bovine spongiform encephalopathy} and \textit{bse} refer to the same concept. If a system predicts both of them as keyphrases, it commits a redundancy error.\\
\textbf{Infrequency errors} occur when a system fails to identify a keyphrase owing to its infrequent presence in the associated document. Handling infrequency errors is a challenge because state-of-the-art keyphrase extractors rarely predict candidates that appear only once or twice in a document. In the \textit{Mad cow disease} example, the keyphrase extractor fails to identify \textit{export} and \textit{scrapie} as keyphrases, resulting in infrequency errors.

\section{Proposed Model}
The WikiRank algorithm includes three steps: (1) Construct the semantic graph including concepts and candidate keyphrases; (2)(optional) Prune the graph with heuristic to filter out candidates which are likely to be erroneously produced; (3) Generate the best set of keyphrases as output.
\subsection{Graph Construction}
\subsubsection{Automatic Concept Annotation}
This is one of the crucial steps in our paper that connects the plain text with human knowledge, facilitating the understanding of semantics. In this step, we adopt \textit{TAGME} \cite{Ferragina:2010:TOA:1871437.1871689} to obtain the underlying concepts in documents.\\
\textit{TAGME} is a powerful topic annotator. It identifies meaningful sequences of words in a short text and link them to a pertinent Wikipedia page, as shown in Figure \ref{example}. These links add a new topical dimension to the text that enable us to relate, classify or cluster short texts.

\subsubsection{Lexical Unit Selection}
This step is to filter out unnecessary word tokens from the input document and generate a list of potential keywords using heuristics. As reported in \cite{Hulth:2003:IAK:1119355.1119383}, most manually assigned keyphrases turn out to be noun groups. We follow \cite{Wan:2008:CTC:1599081.1599203} and select candidates lexical unit with the following Penn Treebank tags: NN, NNS, NNP, NNPS, and JJ, which are obtained using the Stanford POS tagger \cite{Toutanova:2003:FPT:1073445.1073478}, and then extract the noun groups whose pattern is zero or more adjectives followed by one or more nouns. The pattern can be represented using regular expressions as follows
\[
(JJ) * (NN|NNS|NNP|NNPS) +
\]
where JJ indicates adjectives and various forms of nouns are represented using NN, NNS and NNP .

\subsubsection{Graph building} 
\begin{figure}[htb]
\begin{center}
\includegraphics[width = \columnwidth]{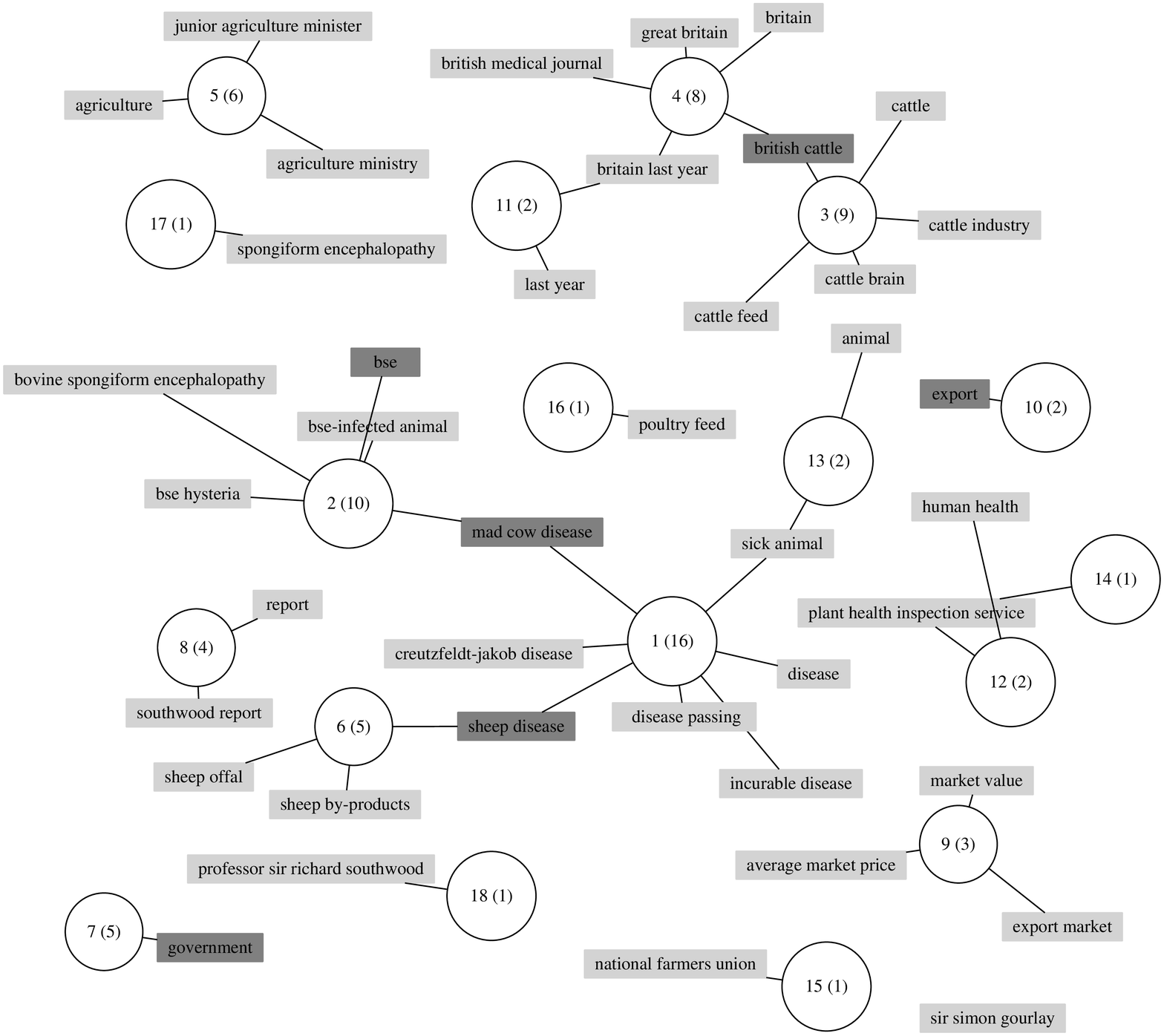} 
\caption[]{\tabular[t]{@{}l@{}} Part of the Semantic Graph of the Sample Document \\\scriptsize Circle: Concept\hspace{10pt} Rectangle: Candidate Keyphrase \hspace{10pt}\\\scriptsize Dark Rectangle: Gold Keyphrase\endtabular}
\label{SemGraph}
\end{center}
\end{figure}

We build a semantic graph $G=[V; E]$ in which the set of vertices $V$ is the union of the concept set $C$ and the candidate keyphrase set $P$---i.e., $V=P\cup C$. In the graph, each unique concept $c\in C$ or candidate keyphrase $p\in P$ for document $d$ corresponds to a node. The node corresponds to a concept $c$ and the node corresponds to a candidate keyphrase $p$ are connected by an edge $(c,p)\in E$, if the candidate keyphrase $p$ contains concept $c$ according to the annotation of \textit{TAGME}. Part of the semantic graph of the sample document is shown in Figure \ref{SemGraph}. Concepts corresponding to \ref{SemGraph} are shown in Table \ref{ConceptTable}.

\begin{table}[!ht]
\begin{center}
\begin{tabularx}{\columnwidth}{|l|l|X|}
\hline
\#&Concept&Frequency\\\hline
1&Disease&16\\\hline
2&Bovine spongiform encephalopathy&10\\\hline
3&Cattle&9\\\hline
4&Great Britain&8\\\hline
5&United States Department&6\\
&of Agriculture&\\\hline
6&Sheep&5\\\hline
7&Government&5\\\hline
8&Report&4\\\hline
9&Market (economics)&3\\\hline
10&Export&2\\\hline
11&Last year&2\\\hline
12&Health&2\\\hline
13&Animal&2\\\hline
14&Animal and Plant Health&1\\
&Inspection Service&\\\hline
15&National Farmers Union&1\\
&of England and Wales&\\\hline
16&Poultry feed&1\\\hline
17&Transmissible spongiform&1\\
&encephalopathy&\\\hline
18&Professor&1\\\hline
\end{tabularx}
\caption{Part of the Concepts Annotated from the Sample Document}
\label{ConceptTable}
\end{center}
\end{table}

\subsection{WikiRank}
\subsubsection{Optimization Problem}
According to \cite{Liu:2010:AKE:1870658.1870694}, good keyphrases should be relevant to the major topics of the given document, at the same time should also have good coverage of the major topics of the document. Since we represent the topical information with concepts annotated with \textit{TAGME}, the goal of our approach is to find the set $\Omega$ consisting of $k$ keyphrases, to cover concepts (1) as important as possible (2) as much as possible.

Let $w_c$ denote the weight of concept $c\in C$. We compute $w_c$ as the frequency $c$ exists in the whole document $d$. To quantify how good the coverage of a keyphrase set $\Omega$ is, we compute the overall score of the concepts that $\Omega$ contains.  

Consider a subgraph of $G$, $G_{sub}$, which captures all the concepts connected to $\Omega$. In $G_{sub}$, the set of vertices $V_{sub}$ is the union of the candidate keyphrase set $\Omega$, and the set $Adj_{\Omega}$ of concepts that nodes in $\Omega$ connect to. The set of edges $E_{sub}$ of $G_{sub}$ is constructed with the edges connect nodes in $\Omega$ with nodes in $Adj_{\Omega}$.

We set up the score of a concept $c$ in the subgraph $G_{sub}$ as following:
\begin{equation}
\label{eq_b}
S(c) = \displaystyle \sum_{i=0}^{deg(c)}\frac{w_c}{2^i}\\
\end{equation}
where $w_c$ is the weight of $c$ as we defined before, and $deg(c)$ is the degree of $c$ in the subgraph $G_{sub}$. Essentially, $deg(c)$ is equal to the frequency that concept $c$ is annotated in the keyphrase set $\Omega$.

The optimization problem is defined as:
\begin{equation}
\begin{array}{rrclcl}
\displaystyle \max_{\Omega} & \multicolumn{3}{l}{\displaystyle \sum_{c\in Adj_{\Omega}} S(c)} \\
\\
\textrm{s.t.} & G_{sub} & = & [V_{sub};E_{sub}] \\
& V_{sub} & = & \Omega\cup Adj_{\Omega} \\
& E_{sub} & = & \{(c,p)|p\in \Omega,c\in Adj_{\Omega} \}\\
& Adj_{\Omega} & = & \{c|c\in \sum_{p\in \Omega}Adj(p)\}\\
& \left|\Omega \right| & \leq & k \\
\end{array}
\end{equation}
The goal of the optimization problem is to find the candidate keyphrase set $\Omega$, such that the sum of the scores of the concepts annotated from the phrases in $\Omega$ is maximized.

\algnewcommand{\Input}[1]{%
\Statex \textbf{Input:}
\Statex \hspace*{\algorithmicindent}\parbox[t]{.8\linewidth}{\raggedright #1}
}
\algnewcommand{\Output}[1]{%
\Statex \textbf{Output:}
\Statex \hspace*{\algorithmicindent}\parbox[t]{.8\linewidth}{\raggedright #1}
}
\algnewcommand{\Initialize}[1]{%
\Statex \textbf{Initialization:}
\Statex \hspace*{\algorithmicindent}\parbox[t]{.8\linewidth}{\raggedright #1}
}

\begin{algorithm}
    \caption{Keyphrase Generalization}
    \begin{algorithmic}[1]

    \Input{
    $|C|$, $P$, $W=\{w_1,\ldots,w_{|C|}\}$\\
    $k$: \Comment{Size of output keyphrase set}\\
    $M_{|P|\times|C|}$: \Comment{Adjacency matrix}
    }
    
    \Output{
    $\Omega$ \Comment{The set of selected keyphrases}
    }
    \Initialize{
    $\Omega\gets\O$\\
    $S=\{s_1\gets 0,\ldots,s_{|P|}\gets 0\}$
    }
    
   \vspace{10pt}
    
    \While {$|\Omega|<k$}
        \For{$p=1$ to $|P|$}
        \State $s_p \gets 0$
            \For{$c= 1$ to $|C|$}
                \If{$M_{p,c}\neq 0$}
                \State $s_p=s_p+w_c$
                \EndIf
            \EndFor
        \EndFor
        \State $q\gets {\arg\max}_{q=1\ldots|P|}\hspace{3pt} s_q$
        \State $\Omega \gets \Omega\cup\{P_q\}$
        \For{$c= 1$ to $|C|$}
            \If{$M_{q,c}\neq 0$}
            \State $w_c \gets w_c/2$
            \EndIf
        \EndFor
    \EndWhile
    \State \Return{$\Omega$}
    \end{algorithmic}
    \label{Algorithm}
\end{algorithm}

\begin{table*}[htb]
\begin{center}
\begin{tabularx}{\textwidth}{ |l|X|X|l|X|X|l|X|X|l|X|X|l|}
      \hline
      &\multicolumn{3}{c|}{DUC}&\multicolumn{3}{c|}{Inspec}&\multicolumn{3}{c|}{ICSI}&\multicolumn{3}{c|}{Nus}\\ \hline
    &P&R&F score&P&R&F score&P&R&F score&P&R&F score\\
      \hline
      SingleRank&26.21&24.45&25.30&25.21&24.10&24.64&3.42&2.49&2.88&0.23&0.98&0.37\\
      \hline
      Topical PageRank&27.33&23.92&25.51&25.58&24.31&24.93&3.98&2.68&3.20&0.64&1.38&0.87\\
      \hline
      Our System&28.72&26.44&27.53&28.14&25.97&27.01&4.71&3.96&4.30&7.27&12.16&9.10\\
      \hline
\end{tabularx}
\caption{The Result of our System as well as the Reimplementation of SingleRank and Topical PageRank on four Corpora}
\label{table.result}
\end{center}
\end{table*}
\subsubsection{Algorithm}
We propose an algorithm to solve the optimization problem, as shown in Algorithm \ref{Algorithm}.
In each iteration, we compute the score $s_p$ for all candidate keyphrases $p\in |P|$ and include the $p$ with highest score into $\Omega$, in which $s_p$ evaluates the score of concepts added to the new set $\Omega$ by adding $p$ into $\Omega$.

\subsection{Approximation Approach with Pre-pruning}
In practice, computing score for all the candidate keyphrases is not always necessary, because some of the candidates are very unlikely to be gold keyphrase that we can remove them from our graph before applying the algorithm to reduce the complexity.

In this section, we introduce three heuristic pruning steps that significantly reduces the complexity of the optimization problem without reducing much of the accuracy.

\textbf{Step 1. Remove the candidate keyphrase $p$ from original graph $G$, if it is not connected to any concept.}

The intuition behind this heuristic is straightforward. Since our objective function is constructed over concepts, if a candidate keyphrase $p$ doesn't contain any concept, adding it to $\Omega$ doesn't bring any improvement to the objective function, so $p$ is irrelevant to our optimization process. Pruning $p$ would be a wise decision.

\textbf{Step 2. Remove the candidate keyphrase $p$ from original graph $G$, if it is only connected to one concept that only exists once in the document}

If a candidate keyphrase contains fewer concepts, or the concepts connects to it barely exist in the document, we think this candidate keyphrase contributes less valuable information to the document. In practice, there are numerous $(c,p)$ pairs in graph $G$ that is isolated from the center of the graph. We believe they are irrelevant to the major topic of the document.

\textbf{Step 3. For a concept $c$ connecting to more than $m$ candidate keyphrases, remove any candidate keyphrase $p\in Adj(c)$ which} (1)Does not connect to any other concept. AND (2)The ranking is lower than $m$th among all candidate keyphrases connect to $c$.(In practice, $m$ is usually 3 or 4.)

According to equation \ref{eq_b}, if there are already $m$ instances of concept $c$ in the $G_{sub}$, adding the $m+1$th instance of $c$ will only contribute $\frac{w_c}{2^m}$ to $S(c)$. At the same time, among all the candidate keyphrases connected to concept $c$, our optimization process always chooses the ones that connect to other concepts as well over the ones that do not connect to any other concept. Combining these two logic, a candidate satisfying the constrains of Step 3 is not likely to be picked in the best keyphrase set $\Omega$, so we can prune it before the optimalization process.

\section{Experiments and Results}

\subsection{Corpora}
The \textbf{DUC-2001} dataset \cite{Over2001}, which is a collection of 308 news articles, is annotated by \cite{Wan:2008:SDK:1620163.1620205}.

The \textbf{Inspec} dataset is a collection of 2,000 abstracts from journal papers including the paper title. This is a relatively popular dataset for automatic keyphrase extraction, as it was first used by \cite{Hulth:2003:IAK:1119355.1119383} and later by Mihalcea and \cite{Mihalcea2004} and \cite{Liu:2009:CFE:1699510.1699544}. 

The \textbf{NUS Keyphrase Corpus} \cite{Nguyen:2007:KES:1780653.1780707} includes 211 scientific conference papers with lengths between 4 to 12 pages. Each paper has one or more sets of keyphrases assigned by its authors and other annotators. The number of candidate keyphrases that can be extracted is potentially large, making this corpus the most challenging of the four.

Finally, the \textbf{ICSI Meeting Corpus} (Janin et al., 2003), which is annotated by Liu et al. (2009a), includes 161 meeting transcriptions. Unlike the other three datasets, the gold standard keys for the ICSI corpus are mostly unigrams.

\subsection{Result}
For comparing with our system, we reimplemented SingleRank and Topical PageRank. Table \ref{table.result} shows the result of our reimplementation of SingleRank and Topical PageRank, as well as the result of our system. Note that we predict the same number of phrase ($k = 10$) for each document while testing all three methods.

The result shows our result has guaranteed improvement over SingleRank and Topical PageRank on all four corpora.

\section{Conclusion and Future Work}
We proposed an unsupervised graph-based keyphrase extraction method WikiRank. This method connects the text with concepts in Wikipedia, thus incorporate the background information into the semantic graph and finally construct a set of keyphrase that has optimal coverage of the concepts of the document. Experiment results show the method outperforms two related keyphrase extraction methods.\\
We suggest that future work could incorporate more other semantic approaches to investigate keyphrase extraction task. Introducing the results of dependency parsing or semantic parsing (e.g., OntoUSP) in intermediate steps could be helpful.


\section{Bibliographical References}
\bibliographystyle{lrec}
\bibliography{lrec}

\end{document}